\documentclass[sigconf]{custom-arxiv}  

\usepackage{booktabs}
\usepackage{paralist}

\usepackage{flushend}
\acmDOI{doi}  
\acmISBN{}  
\acmConference[]{}{}{}
\acmYear{2020}  \copyrightyear{2020}  \acmPrice{}  

\begin{document}

\title{Multi-Vehicle Mixed Reality Reinforcement Learning for Autonomous Multi-Lane Driving}



\author{Rupert Mitchell, Jenny Fletcher, Jacopo Panerati, and Amanda Prorok}
\orcid{}
\affiliation{%
 \institution{Department of Computer Science and Technology, University of Cambridge}
 \streetaddress{William Gates Building, 15 JJ Thomson Ave}
 \city{Cambridge} 
 \state{United Kingdom} 
 \postcode{CB3 0FD}
 \\ \{rmjm3, jlf60, jp872, asp45\}@cam.ac.uk
}

\renewcommand{\shortauthors}{R. Mitchell, J. Fletcher, J. Panerati, and A. Prorok}

\begin{abstract}  
Autonomous driving promises to transform road transport. Multi-vehicle and multi-lane scenarios, however, present unique challenges due to constrained navigation and unpredictable vehicle interactions. 
Learning-based methods---such as deep reinforcement learning---are emerging as a promising approach to automatically design intelligent driving policies that can cope with these challenges. Yet, the process of \textit{safely learning} multi-vehicle driving behaviours is hard: while collisions---and their near-avoidance---are essential to the learning process, directly executing immature policies on autonomous vehicles raises considerable safety concerns.
In this article, we present a safe and efficient framework that enables the learning of driving policies for autonomous vehicles operating in a shared workspace, where the absence of collisions cannot be guaranteed. Key to our learning procedure is a sim2real approach that uses real-world online policy adaptation in a \textit{mixed reality setup}, where other vehicles and static obstacles exist in the virtual domain. This allows us to perform safe learning by simulating (and learning from) collisions between the learning agent(s) and other objects in virtual reality. Our results demonstrate that, after only a few runs in mixed reality, collisions are significantly reduced.
\end{abstract}

\keywords{Multi-robot systems; Machine learning for robotics; Reinforcement learning; Autonomous vehicles; Reality gap; Sim2real}  

\maketitle

\section{Introduction}
\label{sec:intro}

The deployment of automated and autonomous vehicles presents us with transformational opportunities for road transport. To date, the number of companies working on this technology is substantive, and growing~\cite{cbsreport}. 
Opportunities reach beyond single-vehicle automation: by enabling groups of vehicles to jointly agree on maneuvers and navigation strategies, real-time coordination promises to improve overall traffic throughput, road capacity, and passenger safety~\cite{dressler:2014,ferreira2010self}. However, driving in multi-vehicle and multi-lane settings still remains a challenging research problem, due to unpredictable vehicle interactions (e.g., non-cooperative cars, unreliable communication), hard workspace limitations (e.g., lane topographies), and constrained platform dynamics (e.g., steering kinematics, driver comfort).

Learning-based methods, such as deep reinforcement learning, have proven effective at designing robot control policies for an increasing number of tasks in single-vehicle systems, for applications such as navigation~\cite{khan2019learning}, flight~\cite{molchanovSimto2019}, and locomotion~\cite{tan2018sim}. 
Leveraging such methods for learning autonomous driving policies is emerging as a particularly promising approach~\cite{pan2017virtual, shalev2016safe, kuderer2015learning}.
Yet, the process of \textit{safely learning} autonomous driving involves unique challenges, since the decision models often used in robotics do not lend themselves naturally to the multi-vehicle domain, due to the unpredictable behaviour of other agents. The unapologetic nature of the trial-and-error process in reinforcement learning compounds the difficulty of ensuring functional safety.

These adversities call for learning that first takes place in simulation, before transferring to the real world~\cite{miglino1995evolving, shah2018airsim}. 
This transfer, often referred to as \textit{sim2real}, is challenging due to discrepancies between conditions in simulation and the real world (such as vehicle dynamics and sensor data)~\cite{peng2018sim, james2019sim, chebotar2019closing}.
Despite substantial advances in this field, the problem of executing immature policies directly on an autonomous vehicle still raises considerable safety concerns. These concerns are exacerbated when \textit{multiple autonomous vehicles share the same workspace}, risking collisions and un-reparable damage.
Simultaneously, the act of colliding---or nearly-colliding---is essential to the learning process, enabling future policy roll-outs to incorporate these critical experiences. \textit{How are we to provide safe multi-vehicle learning experiences, without forgoing the realism of high-fidelity training data?}
There is a dearth of work that addresses this challenge.

\begin{figure}
	\hspace*{-0.3cm}\includegraphics[]{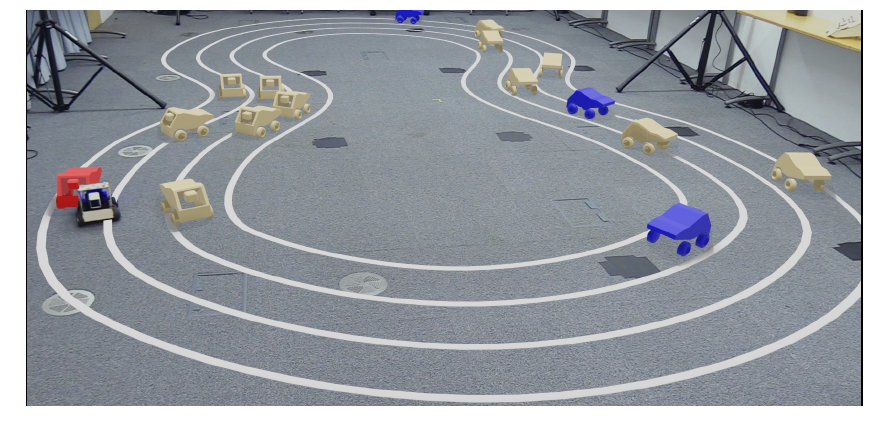}
	\caption{Mixed reality multi-vehicle multi-lane traffic circuit including one real DeepRacer robot and twelve virtual ones, in beige. Four static virtual vehicles are rendered in blue. The colliding virtual vehicle is rendered in red.}
	\label{fig:setup}
\end{figure}

Our goal in this paper is to develop a safe and efficient framework that allows us to learn driving policies for autonomous vehicles operating in a shared workspace, where collision-freeness cannot be guaranteed.
Rather than focusing on re-elaborating or advancing state-of-the-art reinforcement learning, our desire is to make it directly applicable onto physical robots. 
Towards this end, we learn an end-to-end policy for vehicle navigation on a multi-lane track that is shared with other moving vehicles and static obstacles. 
The learning is based on a model-free method embedded in a distributed training mechanism that we tailor for \textit{mixed reality} compatibility. 
Key to our learning procedure is a sim2real approach that uses real-world online policy adaptation in a \textit{mixed reality setup}, where obstacles (vehicles and objects) exist in the virtual domain. This allows us to perform safe learning by simulating (and learning from) collisions between the learning agent(s) and other objects in virtual reality. 
We apply our framework to a multi-vehicle setup consisting of one real vehicle, and several simulated vehicles (as shown in Figure~\ref{fig:setup}). Experiments show that a significant performance improvement can be obtained after just a few runs in mixed reality, reducing the number of collisions and increasing reward collection.
To the best of our knowledge, this is the first demonstration of mixed reality reinforcement learning for multi-vehicle applications.

\section{Related Work}
\label{sec:related}

Training in simulation before transferring learned policies to the real world provides the benefits of safety and facilitated data collection. Several methods alleviate the difficulty of bridging the reality gap: \emph{(i)} parameter estimation, which estimates parameters of the real system to achieve a more realistic simulation~\cite{lowrey2018reinforcement, tan2018sim}, \emph{(ii)} iterative data collection, which learns distributions of dynamics parameters in an iterative manner~\cite{christiano2016transfer, chebotar2019closing}, and \emph{(iii)} domain randomization, which trains over a distribution of the system dynamics for policies that are more robust against simulator discrepancies from reality~\cite{peng2018sim, muratore2018domain, james2019sim, tobin_Domain_2017}. Although these methods contribute significantly to closing the reality gap, the problem of guaranteeing safe policy execution still persists. 
Moreover, it often proves hard to accommodate all situations the robot may encounter in the real world, where unexpected conditions are the norm. To ease this challenge, researchers have proposed methods for continuous online adaptation in model-based reinforcement learning~\cite{fu2016one, gu2016continuous}. The aim of this approach is to learn an approximate model and then adapt it at test time. However, this can still lead to safety concerns when there is a mismatch between what the model is trained for, and how it is used at test-time. More recent approaches, such as meta-learning, strive to overcome this challenge~\cite{nagabandi_Learning_2019}. The commonality of all these approaches, however, is their focus on single-robot systems in isolated work-spaces; guaranteeing safe online-learning in shared workspaces is still an open problem.

The idea of exploiting mixed (and augmented) reality for robotics applications was originally introduced as a tool to facilitate development and prototyping. Early work experiments with virtual humanoids amongst real obstacles~\cite{stilman2005augmented}, leveraging the setup to rapidly prototype and test humanoid sub-components. Chen et al. \cite{chen2009mixed} use augmented reality to obtain a coherent display of visual feedback during interactions between a real robot and virtual objects. More recently, mixed reality has gained importance in shared human-robot environments~\cite{williams2018virtual}, where combinations of physical and virtual environments can provide safer ways to test interactions, \textit{``... by also allowing a gradual
transition of the system components into shared physical environments''}~\cite{hoenig2015mixed}.
The introduction of mixed reality to support reinforcement learning has barely been considered. In~\cite{mohammadi2019mixed}, Mohammadi et al. present an approach for online continuous deep reinforcement learning for a reach-to-grasp task in a mixed reality environment. Although targets exist in the physical world, the learning procedure is carried out in simulation (using real data), before actions are transferred and executed on the actual robot.

The particularity of our work is that we focus on multi-robot settings, where inter-robot interactions contribute significantly to the learning process, but cannot be executed directly on multiple real platforms without incurring repeated damages.
Not only does our mixed reality framework help bridge the reality gap that still stymies progress in reinforcement learning for robotics, but also, it is especially significant for the specific application at hand in this work.

\section{Problem Statement}
\label{sec:problem}

We consider a multi-vehicle system composed of $N$ vehicles on a multi-lane (closed) traffic circuit with $M$ lanes. Each vehicle in the system has a unique target velocity, $v_t$, i.e., vehicles aim to travel at potentially different speeds. The circuit is obstructed by $K$ obstacles (static vehicles).
In order to maintain target speeds and avoid collisions, vehicles must learn to change lanes and execute overtaking maneuvers (we do not enforce a rule regarding which side a vehicle may overtake on). An image of our three-lane setup is shown in Figure~\ref{fig:setup}, with 13 vehicles (one of which is real) and 4 virtual obstacles (in blue).

\textbf{Assumptions.}
We are especially interested in a vehicle's high-level decision-making process that involves lane changes and speed modulation. We, therefore, consider the availability of a low-level controller that executes reliable trajectory following, allowing the vehicle to remain in the centre of its current lane.
To facilitate the low-level control task, we represent a lane by a sequence of cubic Bezier curves, continuous up to their first derivative (i.e. having no sharp corners). Vehicles are provided reliable positioning information (e.g., through a motion capture system).
We also assume the ability of basic local communication, such that the desired velocity of each neighbouring vehicle is available to the high-level controller. This neighbourhood includes the six nearest vehicles within a vision radius, $r_v$.
Our vehicles' knowledge is thus local.
We do not directly deal with noisy perception, as our sim2real challenge is the result of non-ideal vehicle models. 
We observe, however, that imperfect sensing would exacerbate this, and our work would prove equally or more valuable in such scenarios.

\textbf{Goal.} Our goal is to \textit{learn a high-level control policy} that allows  vehicles to drive as closely as possible to their target velocities, while avoiding collisions with other vehicles.

\section{Multi-Vehicle System}
\label{sec:vehicles}

Our multi-vehicle system is based on a physical vehicle, the DeepRacer robot~\cite{balaji2019deepracer}, for which we also develop a virtual counterpart. 
This platform, its dynamics, and control model are detailed below.

\subsection{The DeepRacer Robot}
The DeepRacer is a 1/18th scale car with a 4MP camera, 4-wheel drive and Ackermann steering. It sports an Intel Atom processor, 4GB of memory, and 32GB of storage. It runs Ubuntu 16.04 LTS and ROS Kinetic Kame. The on-board computer and motors are powered by 13600mAh and 1100mAh batteries, respectively.

The DeepRacer was originally designed as a platform for vision-based reinforcement learning, with training carried out in simulation only. 
This is different to our aim---which includes online training and but also only focuses on non-vision-based, high-level decision-making. Therefore, we modified the platform to make it more suited to our goal.
The default ROS launch script was replaced, so that the DeepRacer does not
run a ROS master but relies on one running on a different device---therefore allowing more than one DeepRacer to be controlled simultaneously. 
We implemented a new ROS node to communicate with the DeepRacer's servo node to set turning and throttle values. Adding this node also meant that communication to the DeepRacer could be done via UDP, reducing latency. 
Finally,
a custom, non-reflective case was designed to allow the integration of the robot with a motion tracking system.

\subsection{Vehicle Model}
The DeepRacer has Ackermann steering geometry. We approximate its kinematics by the bicycle model, with motion equations:
\begin{eqnarray} 
\label{eq:equation_of_motion}
\dot{x} &=& v_c\cos{\xi} \nonumber \\
\dot{y} &=& v_c\sin{\xi} \nonumber \\
\dot{\psi} &=& L^{-1}{v_c\tan{\phi_s}},
\end{eqnarray}
where $\phi_s$ is the steering angle, $v_c$ is the forward speed, $\xi$ is the heading, and $L$ is the vehicle's wheel base.
These equations are numerically integrated in our simulation \emph{via} the Euler method to obtain the position of the DeepRacer at each time step.
For the purpose of collision detection in mixed reality, the DeepRacer was modeled by a bounding box of similar size to its physical dimensions ($\sim 30cm \times 20cm$). 
Virtual vehicles are also identically modeled.

\subsection{Two-Level Driving Strategy}
\label{sec:control_levels}
We segregate the vehicle's driving strategy into two levels: a high-level controller that is responsible for \emph{(i)} lane-change decisions and \emph{(ii)} velocity modulation, and a low-level controller that acts upon this information to track desired lanes at desired speeds. \textit{In Section~\ref{sec:learning}, our objective is to learn the high-level control policy only.} 
We assume the existence of background traffic that is deployed with a fixed high-level driving strategy.

\textbf{Low-level control.}
Two low-level controllers are used for lateral and longitudinal control. A PID controller onboard the DeepRacer maintains the robot's forwards velocity at the value requested by the high-level controller. The steering angle $\phi_s$ of the DeepRacer is set by a PD controller, keeping the robot on the trajectory chosen by the higher level controller.
The onboard velocity controller gets a desired velocity $v_c$ from the high-level controller, and pose information from the motion tracking system; it calculates velocity and acceleration towards the desired trajectory. These are used in the PID controller which outputs a throttle value to the motors. This allows the DeepRacer to travel at the speed requested by the high level controller regardless of external factors such as how discharged the battery is.

The objective of the steering angle controller is to minimise the perpendicular distance, $\delta$, between the robot and the desired trajectory. For small deviations, the angle of the robot's heading with respect to the trajectory, $\psi$, is proportional to $\frac{d \delta}{d s}$ and the steering angle of the robot, $\phi_s$, is proportional to $\frac{d^2 \delta}{d s^2}$, where $s$ is the traveled distance. This permits a controller of the form $\phi_s=-g\delta-gd\tan\psi+l\kappa$, where $\kappa$ is the curvature of the trajectory at the nearest point and $g$ and $d$ are gain and damping factors, respectively. The use of $\tan\psi$ in place of $\psi$
causes the robot to continue to converge to the desired trajectory even for larger deviations,
not affecting its behaviour
for small deviations. Since the controller uses derivatives with respect to $s$ rather than $t$ directly, it behaves the same independently of how the high-level controller changes the robot's speed. 

\textbf{High-level control policy.}
While low-level controller is capable of maintaining a specified velocity and following the centre of a chosen lane, we use a high-level control algorithm to decide when to accelerate or decelerate and when to change lanes. This high-level policy is the \textit{learnable policy} (described in Section~\ref{sec:nn}) applied to the agent vehicle. 

\textbf{Background traffic.}
For realistic (virtual) background traffic we use a hard-coded algorithm, following the work in~\cite{hyldmar_Fleet_2019}. This controller has both longitudinal and lateral control components. 
The longitudinal component is based on the Intelligent Driver Model (IDM) proposed in~\cite{treiber2000congested}.
Using this control method, a vehicle's forward acceleration is a function of its current velocity, $v_c$, its gap $s$ to the vehicle in front, and the rate at which it is approaching the vehicle in front, $\Delta v$:
\begin{equation}
    a_{\mathrm{IDM}} = \alpha \left [  1- \left ( \frac{v_c}{v_t} \right)^\delta - \left (\frac{s^{\star}(v_c,\Delta v)}{s} \right )^2  \right ],
\end{equation}
where $s^{\star}$ is a function determining the desired minimum gap to the preceding vehicle and $v_t$ is a target velocity. This gap is defined as:
\begin{equation}
    s^{\star}(v_c,\Delta v) = s_0 + T v_c + \frac{v_c \Delta v}{2 \sqrt{\alpha \beta}},
\end{equation}
where $T$, $\alpha$, $\beta$, $s_0$, $v_t$ are parameters and $s_0$ is a jam distance---the distance which cars in a queue will leave between each other. 

The lateral component of this high level controller, responsible for lane changes, is based on the MOBIL controller proposed in~\cite{kesting2007general}. 
The MOBIL strategy is designed to maximise the current vehicle's freedom to accelerate while also considering the interests of nearby vehicles, and maintaining safety. To determine the effect of a lane change on the current vehicle's own acceleration, the MOBIL controller considers the effect ($\Delta a_{self}$) the new gap to the next vehicle would have on the chosen acceleration by its longitudinal control algorithm, IDM. The MOBIL controller similarly calculates the effect a proposed lane change would have on the chosen accelerations of nearby vehicles, assuming they were also using IDM. It then compares the expected benefit to a threshold value $\Delta a_T$ to determine whether or not to change lane:
\begin{equation}
\Delta a_{self} + p (\Delta a_n + \Delta a_o) > \Delta a_{T},
\end{equation}
where $\Delta a_{n}$ and $\Delta a_o$ are the effects on the new and old following vehicles, and $p$ is a politeness factor. Safety is maintained by adding the condition that the MOBIL controller does not force the new follower vehicle to decelerate at a rate greater than a safety limit, $\beta_n$. Since we do not enforce a rule regarding which side vehicles may overtake on, the MOBIL controller considers changing lanes in both directions, and takes the better option if both surpass the threshold $\Delta a_T$.

\section{Learning Framework}
\label{sec:learning}

As anticipated in Section~\ref{sec:control_levels}, we wish to learn a high-level control policy letting a vehicle avoid collisions while maintaining its desired velocity. We formulate this as a sequential decision problem and solve it with an actor-critic based reinforcement learning approach. We approximate the value function $V$ and the policy function $\pi$ using the critic and actor components, respectively. 
Our implementation is largely inspired by existing literature~\cite{sutton2011,td3,a3c} as our goal is not to advance these techniques, but rather to evaluate their effectiveness in our mixed reality framework.

\subsection{Reinforcement Learning Problem}
\label{sec:rl}

Our goal is to safely (collision and damage-free) find an optimal high-level controller, such that each vehicle (agent) is as close as possible to its desired velocity.
We formalise this high-level control problem as a reinforcement learning problem~\cite{sutton2011} with state space, $\mathcal{O}$ (the agent's observations), and action space $\mathcal{A}$. $\mathcal{O}$ contains both information about the agent's own state, $\mathcal{O}_s$, as well as the state of other nearby vehicles, $\mathcal{O}_o$, such that:
\begin{equation}
    \mathcal{O} = \mathcal{O}_s \times \mathcal{O}_o.
\end{equation}

In $\mathcal{O}_s$, an agent observes:
\emph{(i)} its current velocity, $v_c$;
\emph{(ii)} its target velocity, $v_t$;
\emph{(iii)} the number of lanes to its right, $l_r$;
\emph{(iv)} the number of lanes to its left, $l_l$;
\emph{(v)} its lane-changing state $s$ (i.e. whether it is changing lane or not).
An element of $\mathcal{O}_s$ is thus represented as a vector of the form:
\begin{equation}
    \textbf{o}_s = [v_c, v_t, l_r, l_l, s] \in \mathbb{R}^5.
\end{equation}

In $\mathcal{O}_o$, the agent observes up to six nearby vehicles (defining its neighbourhood, as introduced in Section~\ref{sec:problem}). If there are less than six vehicles within radius $r_v$, then this vector is padded up to six using ``null'' vehicles.
For each nearby vehicle, $c_i$, the agent receives the relative position of $c_i$ in polar coordinates ($d_i$, $\theta_i$). 
The agent also receives the relative lane-wise velocity, $v_{ri}$, of $c_i$, the number of lanes to $c_i$, $\Delta l_i$, and the lane-changing state of $c_i$, $s_i$. An element of $\mathcal{O}_o$ is thus represented as 6 vectors of the form:
\begin{equation}
    \textbf{o}_{oi} = [d_i, cos\theta_i, sin\theta_i, v_{ri}, \Delta l_i, s_i] \in \mathbb{R}^6.
\end{equation}

\begin{figure}
	\includegraphics[]{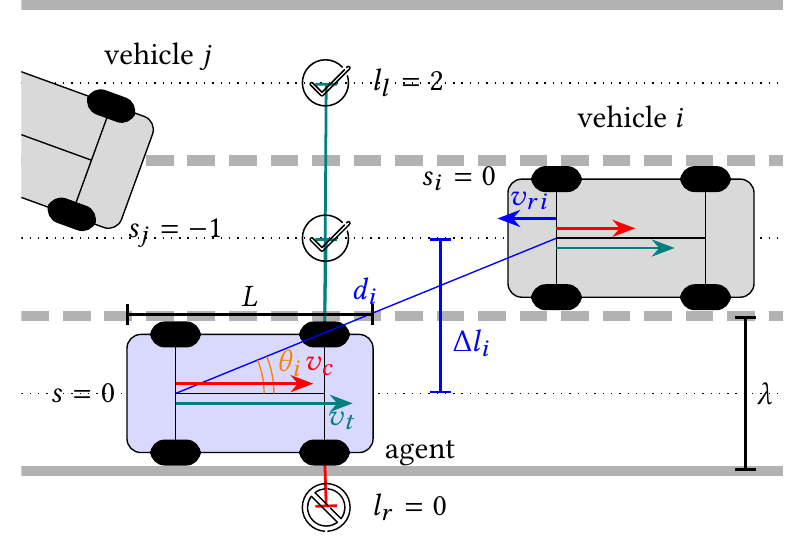}
	\caption{Schematics presenting the main components in the observations vectors $\mathbf{o}_s$ and $\mathbf{o}_{oi}$ for a vehicle tackling the reinforcement learning problem described in Subsection~\ref{sec:rl}.}
	\label{fig:observations}
\end{figure}

The action space, $\mathcal{A}$, contains pairs of tuples from a (discrete) acceleration space, $\mathcal{A}_a$, and a (discrete) lane changing space, $\mathcal{A}_l$, such that:
\begin{equation}
    \mathcal{A} = \mathcal{A}_a \times \mathcal{A}_l.
\end{equation}
Set $\mathcal{A}_a$ consists of ``constant acceleration'', ``maintaining the current speed'', and a ``constant deceleration''. 
Set $\mathcal{A}_l$ consists of ``changing lane left'', ``right'', or ``not at all''.
The reinforcement learning reward function
is designed to prevent the agent from deviating unnecessarily from its desired speed while avoiding collisions with other cars. This function is expressed as:
\begin{equation}
    R(\textbf{o}_s, \textbf{o}_o)=-c_0|v_c-v_t| -\max(p_1, p_2),
\end{equation}
where $p_1$ and $p_2$ are proximity penalty terms defined as:
\begin{equation}
p_1=\max(0, c_1 L - d_l),
\end{equation}
\begin{equation}
p_2=\max(0, c_2 \lambda - d_a),
\end{equation}
where $L$ is the length of a vehicle, $d_l$ is the distance to the closest (ahead or behind) vehicle in the same lane, $\lambda$ is the distance between lanes, $d_a$ is the distance to the closest vehicle (in any lane), and $c_0$, $c_1$, and $c_2$ are parameters (see also Figure~\ref{fig:observations}). 
These two proximity penalties exist to deter the agent from coming too close to other vehicles. While this specific formalization would admit a solution through discrete action-space methods, such as Double Q-learning~\cite{hasselt2010}, in the following, we present a more general approach based on the actor critic method.
As a consequence, our approach can generalise to continuous action spaces as well.

\subsection{Neural Network Architecture}
\label{sec:nn}

\begin{figure}
    \centering
	\includegraphics[scale=1]{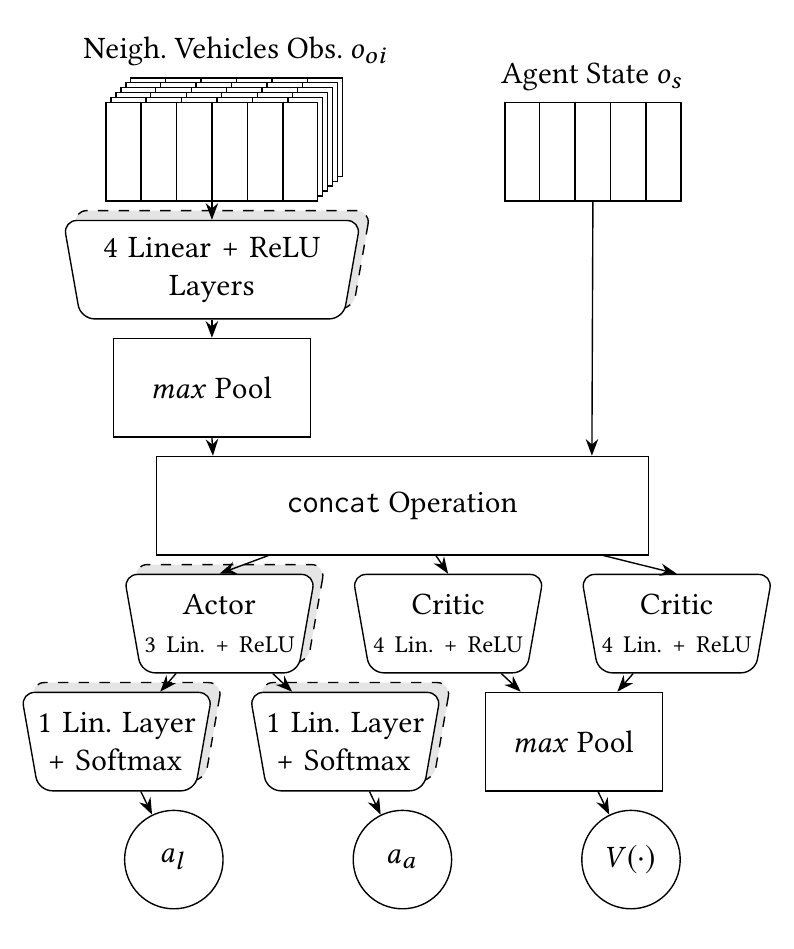}
	\caption{Schematics of the neural network mapping observations $\mathbf{o_s} \in \mathcal{O}_s$, $\mathbf{o_{oi}} \in \mathcal{O}_o$ to \emph{(i)} actions $a_a\in \mathcal{A}_a$, $a_l\in \mathcal{A}_l$ and \emph{(ii)} value function $V(\cdot)$. We detail this architecture in Subsection~\ref{sec:nn} and its training in Subsection~\ref{sec:dist-training}.
	}
	\label{fig:network}
\end{figure}

We approximate value $V(o)$ and policy function $\pi(o,a)$ using a deep neural network containing one actor and two critics (Figure~\ref{fig:network}).
From observation vectors $\mathbf{o}_{oi}$'s, the salient features of nearby cars are extracted using a sequence of four linear layers of hidden size $n_h$ with output size $n_f$. These features are then max-pooled across nearby vehicles to get a single size $n_f$ vector of features pertaining to observed vehicles.
This vector is then concatenated with the agent's own observations $\mathbf{o}_s$ to produce the input of the actor and critic networks.

The actor network consists of a sequence of three linear layers of hidden and output size $n_h$ followed by two heads, each consisting of a final layer of hidden size $n_h$ and an output size of 3, followed by soft-max activation. These two heads correspond to the two discrete spaces $\mathcal{A}_l$ and $\mathcal{A}_a$, i.e., lane changes and acceleration, respectively.
We elect to use two critic networks which are similarly composed by a sequence of four linear layers of hidden size $n_h$, though this time each terminating in a one-dimensional evaluation of the value function.
As proposed by Fujimoto et al.~\cite{td3}, we consider the less extreme of the two evaluations during training to try to reduce the impact of outlier estimations of the value function when updating $\pi$ in the early stages.

\subsection{Distributed Training}
\label{sec:dist-training}

We develop our reinforcement learning method as an adaptation of Asynchronous Advantage Actor Critic (A3C)~\cite{a3c}, by maintaining an approximation for the value function of a state $o$, $V(o)$, and for the policy function $\pi(o,a)$ using explicitly calculated returns over short trajectories. Returns $R_t$ from actions were calculated as
\begin{equation}
R_t=\sum_{i=0}^{k-t}\gamma^ir_{t+i} + \gamma^kV_{avg}(o_{t+k}),
\end{equation}
where $0\leq t<k$ for trajectory length $k$ and $V_{avg}$ is the mean of the two value functions. The approximation of the value function was trained to minimise
$A(o_t,a_t)^2$
where 
$A(o_t,a_t)$ is the Advantage function, $R_t-V(o_t)$.

The policy function is updated using the PPO-Clip~\cite{ppo} loss function:
\begin{equation}
L(\phi)=-\min\left(\rho_t A_\phi(o_t, a_t), f_c(\rho_t, 1- \epsilon, 1 + \epsilon) A_\phi(o_t, a_t)\right),
\end{equation}
where $\phi$ are the network parameters, subscript $\phi$ denotes the evaluation of the network using parameters $\phi$, $f_c$ is the clamp function and $\epsilon$ is a constant parameter:
\begin{equation}
\rho_t = \frac{\pi_\phi(o_t, a_t)}{\pi_{\bar{\phi}}(o_t, a_t)}.
\end{equation}
As we do not use mini-batching, the target policy that we compare against is not one computed before a current set of mini-batches (as in~\cite{ppo}), but rather duplicated versions of part of the network (the shaded boxes in Figure \ref{fig:network}) with parameters smoothed exponentially in time, $\bar{\mathbf{\phi}}$, updated to follow the latest parameters, $\mathbf{\phi}$, according to Polyak-Ruppert averaging:
\begin{equation}
    \bar{\mathbf{\phi}}_{t+1} = \tau \bar{\mathbf{\phi}}_t + (1 - \tau) \mathbf{\phi}_t,
\end{equation}
where $\tau$ is a parameter set during training.
We also add to the loss function a term proportional to the negation of the policy entropy, in order to discourage premature convergence. We weight the three contributions to the total network loss with coefficients $w_a$, $w_c$ and $w_e$ corresponding to the PPO loss, the critic loss and the entropy term, respectively.

\begin{figure}
    \centering
	\includegraphics[scale=1]{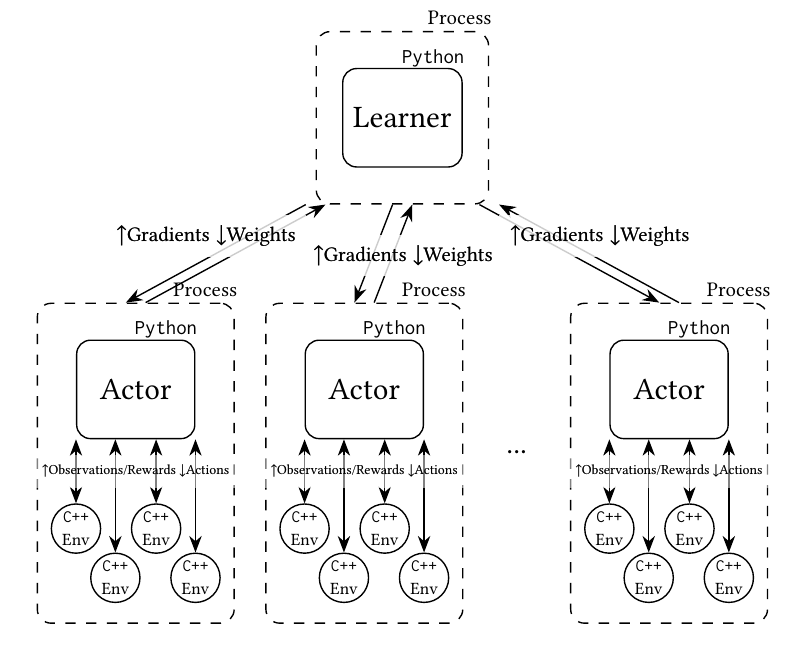}
	\caption{Schematics of the distributed training approach presented in Subsection~\ref{sec:dist-training} for the network in Figure~\ref{fig:network}.}
	\label{fig:parallelism}
\end{figure}

To improve speed and stability of learning, we use multiple parallel actors when pre-training a policy in simulation only. We parallelise this process on two levels. First, we use asynchronous updates, as in \cite{a3c}, to allow multiple threads acting in the problem environment to send gradients to a separate thread updating the policy parameters, and then returning the new parameters (as shown in Figure~\ref{fig:parallelism}). In addition, each actor thread simultaneously acts in multiple environments~\cite{batch-a2c} in order to take advantage of vectorisation (Figure~\ref{fig:parallelism}). Combined, these two parallelisation strategies substantially improved ($\geq$10x speed-up) training speed in purely virtual environments.

\section{Mixed Reality Setup}
\label{sec:setup}

\begin{figure*}
	\includegraphics[]{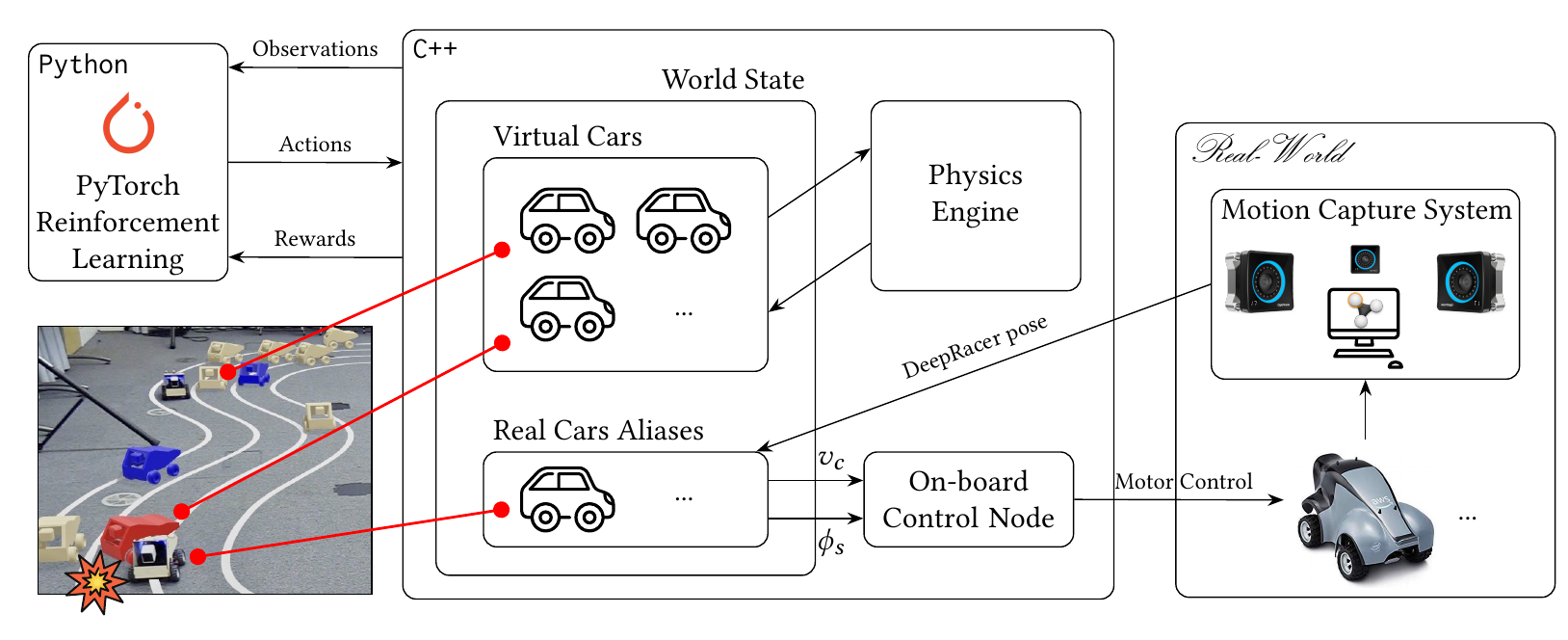}
	\caption{Overall schematics of the proposed multi-vehicle, mixed reality reinforcement learning approach. Reinforcement learning of high-level driving policies is handled through PyTorch. Both virtual and real DeepRacer vehicles exist within a C++ simulation that manages the physics of the virtual cars and emulates collisions in mixed reality. The physics of real-life DeepRacers is captured through OptiTrack's motion capture system and fed to the simulation. }
	\label{fig:approach}
\end{figure*}

Our mixed reality experimental setup seamlessly integrates multiple real-world and virtual components, as illustrated in Figure~\ref{fig:approach}. 
The learning of high-level policies by DeepRacer agents, using the framework presented in Section~\ref{sec:learning}, is performed during the concurrent execution of all these modules, i.e., in mixed reality.

\subsection{Simulation Setup}
In our setup, a C++ simulation provides the environment in which reinforcement learning agents can act, observe, and learn. As such, it also contains the high-level IDM/MOBIL controllers of the background traffic vehicles.
We implemented the reinforcement learning approach described in the previous section using Python and the \texttt{PyTorch} library. An interface between the C++ simulation and the Python interpreter was created using the  \texttt{BOOST.Python} C++ library. This interface exposes the ability to create environments as either \textit{mixed real} or purely virtual. The simulation provides observations and reward signals to the Python implementation, according to the state of the environment. Then, it updates its state to reflect the agents' actions, as received from the Python interpreter.

The simulated environment also contains \emph{(i)} the specifications of the Bezier curves for all lanes in the track, \emph{(ii)} the states of the vehicles controlled by either reinforcement learning agents or the IDM/MOBIL algorithms, and \emph{(iii)} $K$ static obstacles. These obstacles are placed far enough apart to not fully block the road, and so that there is at least one in each lane of the circuit. Their exact positions are otherwise randomised. The starting locations of the background traffic and agent vehicle are likewise randomised along with the desired velocities $v_t$'s of all vehicles.
For each of the vehicles in the environment, collision detection is accomplished using bounding boxes of the same shape and size of a DeepRacer.

The simulation was written in C++ in order to provide higher performance, especially when pre-training a network in a purely simulated environment. 
To the same end, the simulation was designed to be capable of running several simultaneous virtual environments (Figure~\ref{fig:parallelism}) in order to allow the reinforcement learning algorithm to submit multiple parallel actions and receive multiple parallel observations---thus making a more efficient use of our learning computing hardware.

\subsection{Real-World Setup}

As shown in Figure~\ref{fig:approach}, the physical DeepRacer must interface with the simulation while training in mixed reality.
The location and pose of a real-life DeepRacer in the environment is tracked using six OptiTrack Prime 17W cameras and the Motive motion capture software. 
When multiple real DeepRacers are used, we distinguish them by using unique layouts of reflective markers. 
The positions of each of the DeepRacers is broadcast by Motive, received by a \texttt{VRPN client} and published to a ROS topic, making the data available to all nodes in our ROS environment.
In order to reduce network load and increase reliability, the frequency at which poses were transmitted was restricted to 50Hz, since this was also the update rate of the physics engine in the simulation. 
From the perspective of the tracking system, the centre of a vehicle was defined as the centre of its rear axle. This choice preserves consistency with the simulation's definition of the centre of a car---itself chosen for the sake of simplicity, while using an Ackermann steering model.
The vehicles drive on a closed loop track made up of individual trajectories that contain no intersections and are $C^1$ continuous.

\subsection{Mixed Reality}

Mixed reality plays a two-fold role in our work: \textit{(i)} it fosters an agent's learning, allowing simultaneous real and simulated training, and \textit{(ii)} it provides us with better evaluation tools, through the ability to visualise the virtual and real agents' interactions.

\textbf{Learning}
In the mixed reality environment, the simulation receives live updates on the pose of the DeepRacer through the motion capture system and updates its representation of the environment state accordingly. The simulation sends commands setting the steering angle and velocity of the DeepRacer according to the actions of the high-level controller and the lateral component of the low-level controller.

The simulation is able to detect collisions between the DeepRacer and the virtual vehicles through a collision box identical to that of a virtual vehicle sharing the same pose as the real agent. From the point of view of the high-level controllers, including the reinforcement learning agent,
the situation is no different from a purely virtual scenario---with the exception of the world's physics affecting the real DeepRacer. Parallelisation of environments is unavailable when training in a mixed real environment, but since our implementation of A3C uses trajectories of experience with explicitly calculated returns, we substantially increase their length and generate only a small number of trajectories for each optimisation step. Each of these trajectories is created using a different random initialisation of the environment in order to provide a variety of experiences to the reinforcement learning algorithm, at each optimisation step.

\textbf{Visualisation}
To visualise the interaction between the virtual cars and the DeepRacer, during our tests, we set up a fixed camera to record the entire full-length experiments. From the simulation environment, we collect pose data for both the virtual and real cars and compute whether any vehicle is currently experiencing collisions.
These data are processed through a Python script importing Blender's API.
At each timestep, we insert an animation keyframe of a vehicle model in the pose specified by the previously recorded data and a colour determined by whether the vehicle is \textit{(i)} a fixed obstacle (blue), \textit{(ii)} a moving vehicle (beige), or \textit{(iii)}, a vehicle currently in collision (red). 
In a separate scene, the DeepRacer alias is also animated using the same procedure.
These two scenes are then composited together using Z-buffer values so that---when the DeepRacer is in front of a virtual vehicle---the area obscured by the Deepracer is transparent. The output can then be overlayed on top of the test footage to create the effect that the real and virtual vehicles are interacting.
 
\section{Experiments}
\label{sec:experiments}

\begin{figure}
	\includegraphics[]{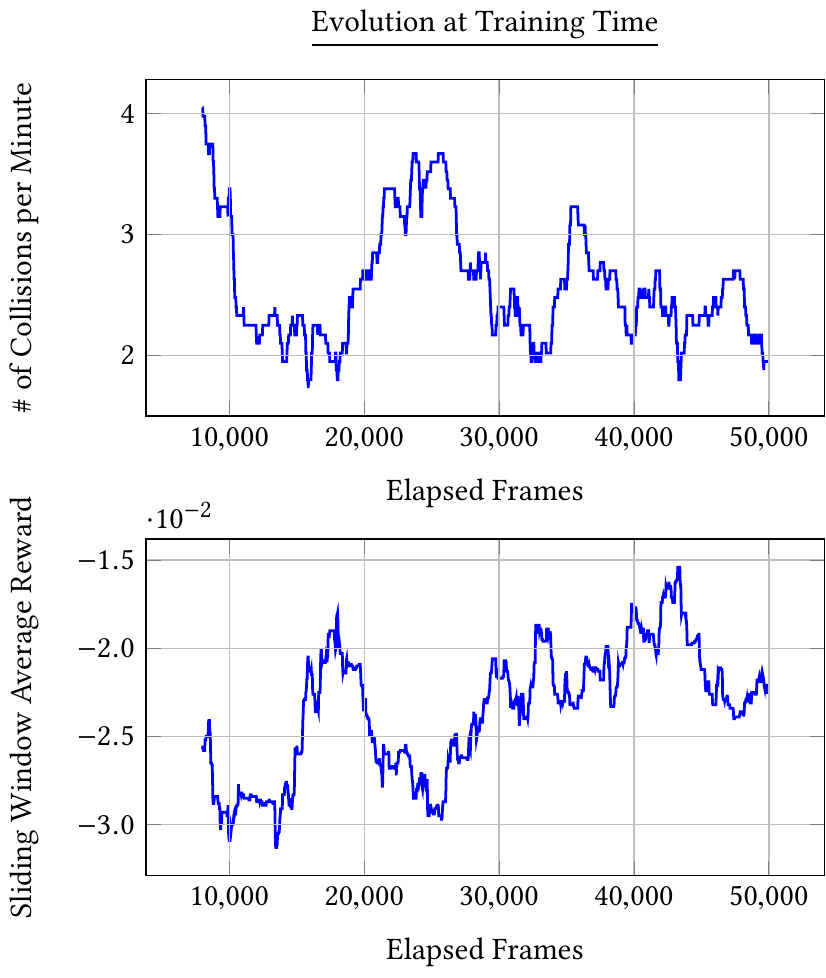}
	\caption{Evolution \underline{during one training instantiation} of \emph{(i)} the number of collisions per minute (top plot, lower is better) and \emph{(ii)} the average reward collected by the training agent, over a sliding window of 8'000 frames (bottom plot, higher is better).}
	\label{fig:evolutions}
\end{figure}

\begin{figure}
	\includegraphics[]{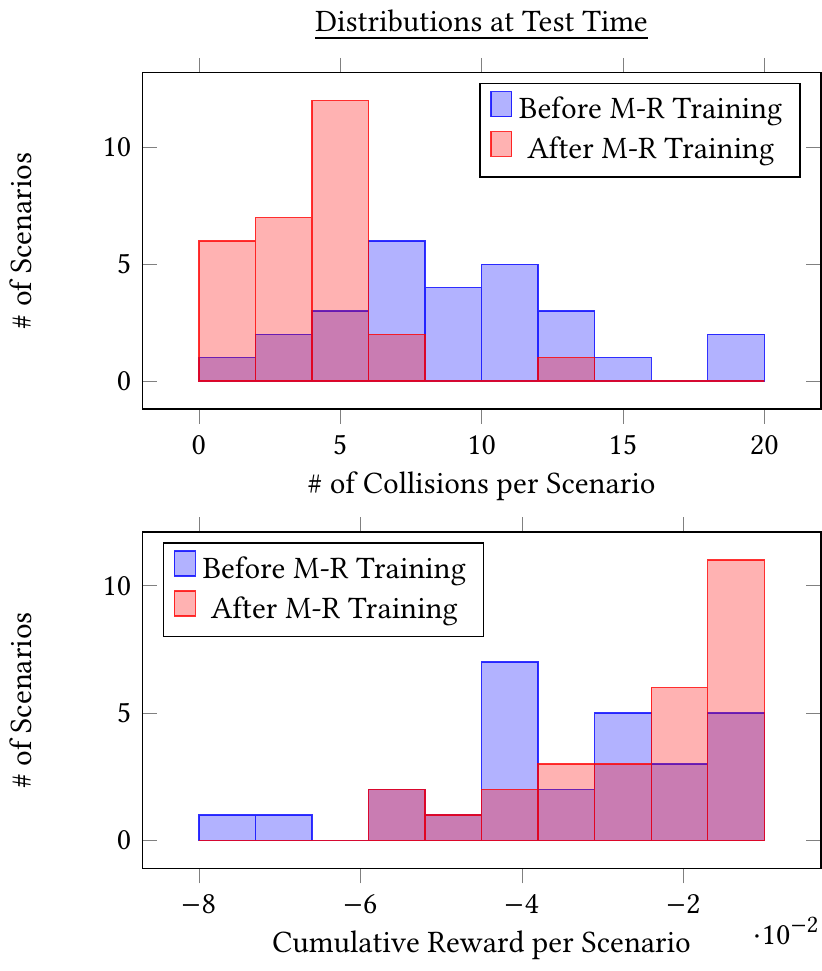}
	\caption{Empirical distributions \underline{at test time} of \emph{(i)} the number of collisions per scenario (top plot, left is best) and \emph{(ii)} the total collected reward per scenario (bottom plot, right is best) before (blue) and after (red) training in mixed reality.}
	\label{fig:distributions}
\end{figure}

\begin{figure}
	\includegraphics[]{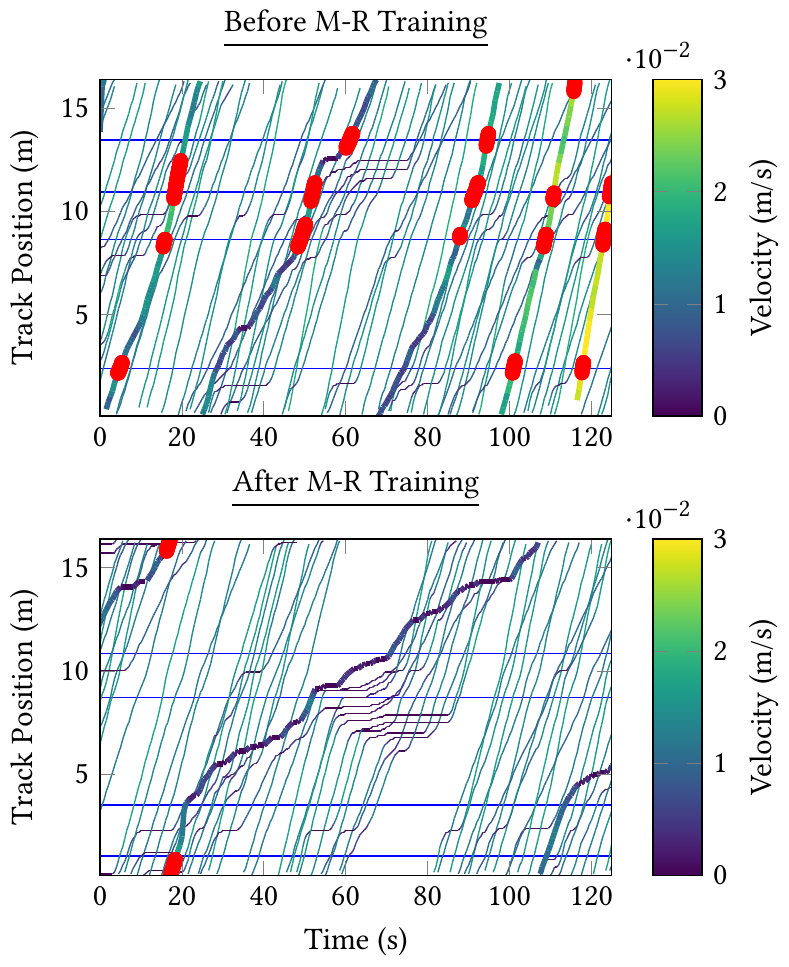}
	\caption{Plots of track positions ($y$ axis) against time ($x$ axis) of four static obstacles (horizontal lines), twelve virtual vehicles, and one real-life DeepRacer (thicker line). The colormap captures the velocities of all cars. The red dots represent collisions incurred by the DeepRacer. The top and bottom plots compare behaviours recorded before and after mixed reality training. 
}
	\label{fig:waves}
\end{figure}

To demonstrate the effectiveness of our mixed reality setup---to train agents capable of collision-free driving---we performed experiments 
on a ($M=$) $3$-lane track (see Figure~\ref{fig:setup}) with lanes $\lambda = 30cm$ wide. The track itself fits a $3.5m \times 2.2m$ area, with a lap length of roughly $16.4$ metres, i.e., $\sim$50 times the size of a DeepRacer ($L=32cm$).
Our experiments include $N=13$ (1 real, 12 virtual) vehicles and $K=4$ virtual obstacles.
The low-level control parameters $g$ and $d$ (see Subsection~\ref{sec:control_levels}) were set to $3$ and $0.4$, respectively.
For the learning parameters (see Section~\ref{sec:learning}),
we selected
$\gamma = 0.9$,
$\tau = 0.7$,
$\epsilon = 0.1$,
$k=128$,
$w_a=10$, $w_c=1$, $w_e=0.003$,
$n_h = 64$,
$n_f = 8$,
$c_0=0.06$, $c_1=0.833$, and $c_2=2.81$.
For the actor and critics, we used learning rates of 2e-4 and 2e-3.
Our results are summarised in Figures~\ref{fig:evolutions}, \ref{fig:distributions}, and \ref{fig:waves} as well as by additional footage available on the \emph{Prorok Lab} YouTube channel.\footnote{\url{https://www.youtube.com/watch?v=LlnaxZHWQOs}}

First, we want to assess the soundness of our approach by evaluating how well training fares---in terms of incurred collisions and collected reward.
This is shown in Figure~\ref{fig:evolutions}, where the two plots describe the evolution over time in a given scenario (measured in frames, i.e., the steps in which an agent receives one set of observations and takes one action) of: \emph{(i)} the number of \emph{collisions per minute} (top plot of Figure~\ref{fig:evolutions}); and \emph{(ii)} the \emph{average collected reward} (bottom plot of Figure~\ref{fig:evolutions}).
Collisions per minute are computed as those yielded by a 8'000-frame long sliding window.
Successful training is reflected in a general downward slope of the top plot (fewer collisions) and, conversely, a general upward slope of the bottom plot (greater reward).

Second, we want to quantify the effectiveness of mixed reality training \emph{at test time}.
This is shown in Figure~\ref{fig:distributions}.
The top and bottom plots refer, once more, to \emph{collisions} and \emph{collected reward}, respectively.
Each one of the two plots compares two density distributions of these performance metrics: one before (in blue) and one after (in red) training in mixed reality.
As our simulation environment is partially randomised, the word \emph{scenario} refers to all the data gathered from a single instantiation.
On the top plot, we can observe a left-shift (from blue to red, i.e., before and after) of the collisions' density distribution, that is, fewer collisions occurring after mixed reality training.
On the bottom plot, conversely, a right-shift reflects the improved ability of the agent, trained in mixed reality, to collect reward.

Finally, Figure~\ref{fig:waves} presents a qualitative comparison of how a DeepRacer agent's behaviour changes before (top) and after (bottom) mixed reality training.
The $x$ axis in Figure~\ref{fig:waves} shows the passing of time (in seconds) while the $y$ axis captures the position of a vehicle along the track (in metres).
Four blue horizontal lines represent obstacles (i.e., static virtual vehicles) on the track.
All other (13) lines represent moving vehicles---the thicker one being the DeepRacer agent.
A colour map is used to encode the speed (in metres per second) of each vehicle.
Red dots indicate collisions between the real-life DeepRacer and either a virtual obstacle or vehicle.
Indeed, collisions are rarer after mixed reality training.
Footage of the mixed reality experiments in Figure~\ref{fig:waves} is also available (\href{https://www.youtube.com/watch?v=LlnaxZHWQOs}{link}).

\section{Discussion}
\label{sec:discussion}

The training stability and effectiveness of the proposed approach is reported in Figure~\ref{fig:evolutions}:
in the top plot, one can observe early improvements---i.e., a reduction---in the number of collisions during training. This is followed by two periods of worsening performance (around frames  20'000 and 30'000), and then a more consistent downward trend (from frame 35'000 on).
The early improvements and performance deterioration (until frame 25'000) may be explained by the choice of hyper-parameters. Our learning rates aimed at aggressive policy changes. That is, an agent would have been, at first, too eager to learn how to overly accelerate---and collect more reward---resulting into more early collisions.
The bottom plot, presenting the collection of reward during training, shows a distinct mirroring ($x$ axis symmetry) of the top plot. This is consistent with what we would expect---that is, a sanity check confirming that a vehicle was led to fewer collisions by seeking higher reward.

Figure~\ref{fig:distributions} demonstrates the performance of our methodology at test time.
In the top plot, we observe that the density distribution of collisions is significantly shifted to the left after mixed reality training---indicating that our learning approach can effectively reduce collisions.
The after-training distribution is also narrower, suggesting reduced variance and uncertainty.
The bottom plot presents the slightly more trivial result that reinforcement learning training does, indeed, lead to improved reward collection. Nonetheless, at test time, this is evidence of the ability of our approach to generalise.

The qualitative results in Figure~\ref{fig:waves} 
demonstrate how the learning agent's behaviour changes before and after mixed reality training.
In the top plot, a DeepRacer that has not yet been trained in mixed reality collides remarkably often, with nearly every obstacle.
This collision-prone behaviour may be due to the reduced responsiveness of the real DeepRacer hardware---when compared to the simulated vehicle---making it harder for the agent to timely stop or avoid other vehicles.
After training in mixed reality, collisions are almost completely amended.
In the bottom plot of Figure~\ref{fig:waves}, we can also observe virtual agents (IDM/MOBIL background traffic) either \emph{(i)} overtaking the learning agent in the longer gaps between obstacles or \emph{(ii)} piling-up behind it in more constrained regions of the road---e.g., when the agent is cautiously approaching two near obstacles. 
Interestingly enough, traffic (e.g., between 50'' and 80'' in the bottom plot of Figure~\ref{fig:waves}) is likely exacerbated by the fact that IDM/MOBIL agents would be willing to give the agent room to accelerate instead of overtaking it---yet, the agent proceeds at a reduced speed. While the learning agent is less dangerous after training, its unexpected prudence can mislead the other driving agents---which are not capable of learning---and reduce throughput.
While the slower speed of the real DeepRacer might appear as a sub-optimal outcome, we should remember that our aim was not to outperform the IDM/MOBIL vehicles---in fact, these can achieve a higher safe speed as their virtual models are only simulated and, thus, more responsive and easier to control than the actual DeepRacers.

Finally, it is important to observe that the simulation performance
of the agents we transferred into our framework was still
characterised by relatively high entropy.
This choice was made to minimise the risk of overfitting to the 
simulation environment and let agents adapt more quickly
to the mixed reality setup.
While we cannot say whether additional simulation-only training
would have benefited or hurt the agents transferring to mixed reality,
our results support the idea that this approach led to quick and effective real-world adaptation.
In future developments of our framework, we will investigate
more sample-efficient off-policy reinforcement learning methods---e.g., \cite{softac} which might allow for better performance without the need for a substantial increase in data gathering---and continuous action spaces.

\section{Conclusions}
\label{sec:conclusions}

This work presented a {mixed reality framework} for safe and efficient reinforcement learning of driving policies in multi-vehicle systems.
Our learning algorithm was trained using a distributed mechanism specifically tailored to suit the needs of our mixed reality setup. 
We demonstrated successful online policy adaptation in an experimental setup involving one real vehicle and sixteen virtual vehicles. Our results showed that mixed reality learning is able to provide significant performance improvements, leading to a reduction of collisions in the learned policies.

The particularity of our system is that it focuses on multi-robot settings, where interactions with other dynamic objects contribute significantly to the learning process, but cannot be executed directly on multiple real platforms without incurring repeated damages.
The proposed framework is a first of its kind: beyond providing specific benefits to the application at hand, it also helps bridge the reality gap that still stymies progress in reinforcement learning for robotics at large.
Future work will consider multiple learning agents using on-board sensing (e.g., vision), and how our mixed reality setup enables their gradual introduction into mutually shared spaces.

\section*{Acknowledgements}
This work was supported by the Engineering and Physical Sciences Research Council (grant EP/S015493/1). Their support is gratefully acknowledged.
The DeepRacer robots used in this work were a gift to Amanda Prorok from AWS. Their support is gratefully acknowledged. This article solely reflects the opinions and conclusions of its authors and not AWS or any other Amazon entity.



\begin{thebibliography}{37}


\ifx \showCODEN    \undefined \def \showCODEN     #1{\unskip}     \fi
\ifx \showDOI      \undefined \def \showDOI       #1{#1}\fi
\ifx \showISBNx    \undefined \def \showISBNx     #1{\unskip}     \fi
\ifx \showISBNxiii \undefined \def \showISBNxiii  #1{\unskip}     \fi
\ifx \showISSN     \undefined \def \showISSN      #1{\unskip}     \fi
\ifx \showLCCN     \undefined \def \showLCCN      #1{\unskip}     \fi
\ifx \shownote     \undefined \def \shownote      #1{#1}          \fi
\ifx \showarticletitle \undefined \def \showarticletitle #1{#1}   \fi
\ifx \showURL      \undefined \def \showURL       {\relax}        \fi
\providecommand\bibfield[2]{#2}
\providecommand\bibinfo[2]{#2}
\providecommand\natexlab[1]{#1}
\providecommand\showeprint[2][]{arXiv:#2}

\bibitem[\protect\citeauthoryear{Balaji, Mallya, Genc, Gupta, Dirac, Khare,
  Roy, Sun, Tao, Townsend, et~al\mbox{.}}{Balaji et~al\mbox{.}}{2019}]%
        {balaji2019deepracer}
\bibfield{author}{\bibinfo{person}{Bharathan Balaji}, \bibinfo{person}{Sunil
  Mallya}, \bibinfo{person}{Sahika Genc}, \bibinfo{person}{Saurabh Gupta},
  \bibinfo{person}{Leo Dirac}, \bibinfo{person}{Vineet Khare},
  \bibinfo{person}{Gourav Roy}, \bibinfo{person}{Tao Sun},
  \bibinfo{person}{Yunzhe Tao}, \bibinfo{person}{Brian Townsend},
  {et~al\mbox{.}}} \bibinfo{year}{2019}\natexlab{}.
\newblock \showarticletitle{DeepRacer: Educational Autonomous Racing Platform
  for Experimentation with Sim2Real Reinforcement Learning}.
\newblock \bibinfo{journal}{\emph{arXiv preprint arXiv:1911.01562}}
  (\bibinfo{year}{2019}).
\newblock


\bibitem[\protect\citeauthoryear{CBS}{CBS}{2018}]%
        {cbsreport}
\bibfield{author}{\bibinfo{person}{CBS}.} \bibinfo{year}{2018}\natexlab{}.
\newblock \bibinfo{title}{{CBS Insights Research Brief}}.
\newblock
  \bibinfo{howpublished}{\url{https://www.cbinsights.com/research/autonomous-driverless-vehicles-corporations-list/}}.
\newblock
\newblock
\shownote{(Accessed August 15, 2018).}


\bibitem[\protect\citeauthoryear{Chebotar, Handa, Makoviychuk, Macklin, Issac,
  Ratliff, and Fox}{Chebotar et~al\mbox{.}}{2019}]%
        {chebotar2019closing}
\bibfield{author}{\bibinfo{person}{Yevgen Chebotar}, \bibinfo{person}{Ankur
  Handa}, \bibinfo{person}{Viktor Makoviychuk}, \bibinfo{person}{Miles
  Macklin}, \bibinfo{person}{Jan Issac}, \bibinfo{person}{Nathan Ratliff},
  {and} \bibinfo{person}{Dieter Fox}.} \bibinfo{year}{2019}\natexlab{}.
\newblock \showarticletitle{Closing the sim-to-real loop: Adapting simulation
  randomization with real world experience}. In \bibinfo{booktitle}{\emph{2019
  International Conference on Robotics and Automation (ICRA)}}. IEEE,
  \bibinfo{pages}{8973--8979}.
\newblock


\bibitem[\protect\citeauthoryear{Chen, MacDonald, and Wunsche}{Chen
  et~al\mbox{.}}{2009}]%
        {chen2009mixed}
\bibfield{author}{\bibinfo{person}{Ian Yen-Hung Chen}, \bibinfo{person}{Bruce
  MacDonald}, {and} \bibinfo{person}{Burkhard Wunsche}.}
  \bibinfo{year}{2009}\natexlab{}.
\newblock \showarticletitle{Mixed reality simulation for mobile robots}. In
  \bibinfo{booktitle}{\emph{2009 IEEE International Conference on Robotics and
  Automation}}. IEEE, \bibinfo{pages}{232--237}.
\newblock


\bibitem[\protect\citeauthoryear{Christiano, Shah, Mordatch, Schneider,
  Blackwell, Tobin, Abbeel, and Zaremba}{Christiano et~al\mbox{.}}{2016}]%
        {christiano2016transfer}
\bibfield{author}{\bibinfo{person}{Paul Christiano}, \bibinfo{person}{Zain
  Shah}, \bibinfo{person}{Igor Mordatch}, \bibinfo{person}{Jonas Schneider},
  \bibinfo{person}{Trevor Blackwell}, \bibinfo{person}{Joshua Tobin},
  \bibinfo{person}{Pieter Abbeel}, {and} \bibinfo{person}{Wojciech Zaremba}.}
  \bibinfo{year}{2016}\natexlab{}.
\newblock \showarticletitle{Transfer from simulation to real world through
  learning deep inverse dynamics model}.
\newblock \bibinfo{journal}{\emph{arXiv preprint arXiv:1610.03518}}
  (\bibinfo{year}{2016}).
\newblock


\bibitem[\protect\citeauthoryear{Clemente, Mart{\'{\i}}nez, and
  Chandra}{Clemente et~al\mbox{.}}{2017}]%
        {batch-a2c}
\bibfield{author}{\bibinfo{person}{Alfredo~V. Clemente},
  \bibinfo{person}{Humberto Nicol{\'{a}}s~Castej{\'{o}}n Mart{\'{\i}}nez},
  {and} \bibinfo{person}{Arjun Chandra}.} \bibinfo{year}{2017}\natexlab{}.
\newblock \showarticletitle{Efficient Parallel Methods for Deep Reinforcement
  Learning}.
\newblock \bibinfo{journal}{\emph{arXiv preprint arXiv:1705.04862}}
  (\bibinfo{year}{2017}).
\newblock


\bibitem[\protect\citeauthoryear{Dressler, Hartenstein, Altintas, and
  Tonguz}{Dressler et~al\mbox{.}}{2014}]%
        {dressler:2014}
\bibfield{author}{\bibinfo{person}{Falko Dressler}, \bibinfo{person}{Hannes
  Hartenstein}, \bibinfo{person}{Onur Altintas}, {and} \bibinfo{person}{Ozan
  Tonguz}.} \bibinfo{year}{2014}\natexlab{}.
\newblock \showarticletitle{Inter-vehicle communication: Quo vadis}.
\newblock \bibinfo{journal}{\emph{IEEE Communications Magazine}}
  \bibinfo{volume}{52}, \bibinfo{number}{6} (\bibinfo{year}{2014}),
  \bibinfo{pages}{170--177}.
\newblock


\bibitem[\protect\citeauthoryear{Ferreira, Fernandes, Concei{\c{c}}{\~a}o,
  Viriyasitavat, and Tonguz}{Ferreira et~al\mbox{.}}{2010}]%
        {ferreira2010self}
\bibfield{author}{\bibinfo{person}{Michel Ferreira}, \bibinfo{person}{Ricardo
  Fernandes}, \bibinfo{person}{Hugo Concei{\c{c}}{\~a}o},
  \bibinfo{person}{Wantanee Viriyasitavat}, {and} \bibinfo{person}{Ozan~K
  Tonguz}.} \bibinfo{year}{2010}\natexlab{}.
\newblock \showarticletitle{Self-organized traffic control}. In
  \bibinfo{booktitle}{\emph{Proceedings of the seventh ACM international
  workshop on VehiculAr InterNETworking}}. ACM, \bibinfo{pages}{85--90}.
\newblock


\bibitem[\protect\citeauthoryear{Fu, Levine, and Abbeel}{Fu
  et~al\mbox{.}}{2016}]%
        {fu2016one}
\bibfield{author}{\bibinfo{person}{Justin Fu}, \bibinfo{person}{Sergey Levine},
  {and} \bibinfo{person}{Pieter Abbeel}.} \bibinfo{year}{2016}\natexlab{}.
\newblock \showarticletitle{One-shot learning of manipulation skills with
  online dynamics adaptation and neural network priors}. In
  \bibinfo{booktitle}{\emph{2016 IEEE/RSJ International Conference on
  Intelligent Robots and Systems (IROS)}}. IEEE, \bibinfo{pages}{4019--4026}.
\newblock


\bibitem[\protect\citeauthoryear{Fujimoto, van Hoof, and Meger}{Fujimoto
  et~al\mbox{.}}{2018}]%
        {td3}
\bibfield{author}{\bibinfo{person}{Scott Fujimoto}, \bibinfo{person}{Herke van
  Hoof}, {and} \bibinfo{person}{David Meger}.} \bibinfo{year}{2018}\natexlab{}.
\newblock \showarticletitle{Addressing Function Approximation Error in
  Actor-Critic Methods}. In \bibinfo{booktitle}{\emph{Proceedings of the 35th
  International Conference on Machine Learning}}
  \emph{(\bibinfo{series}{Proceedings of Machine Learning Research})},
  \bibfield{editor}{\bibinfo{person}{Jennifer Dy} {and}
  \bibinfo{person}{Andreas Krause}} (Eds.), Vol.~\bibinfo{volume}{80}.
  \bibinfo{publisher}{PMLR}, \bibinfo{address}{Stockholmsm\"{a}ssan, Stockholm
  Sweden}, \bibinfo{pages}{1587--1596}.
\newblock
\urldef\tempurl%
\url{http://proceedings.mlr.press/v80/fujimoto18a.html}
\showURL{%
\tempurl}


\bibitem[\protect\citeauthoryear{Gu, Lillicrap, Sutskever, and Levine}{Gu
  et~al\mbox{.}}{2016}]%
        {gu2016continuous}
\bibfield{author}{\bibinfo{person}{Shixiang Gu}, \bibinfo{person}{Timothy
  Lillicrap}, \bibinfo{person}{Ilya Sutskever}, {and} \bibinfo{person}{Sergey
  Levine}.} \bibinfo{year}{2016}\natexlab{}.
\newblock \showarticletitle{Continuous deep q-learning with model-based
  acceleration}. In \bibinfo{booktitle}{\emph{International Conference on
  Machine Learning}}. \bibinfo{pages}{2829--2838}.
\newblock


\bibitem[\protect\citeauthoryear{Haarnoja, Zhou, Abbeel, and Levine}{Haarnoja
  et~al\mbox{.}}{2018}]%
        {softac}
\bibfield{author}{\bibinfo{person}{Tuomas Haarnoja}, \bibinfo{person}{Aurick
  Zhou}, \bibinfo{person}{Pieter Abbeel}, {and} \bibinfo{person}{Sergey
  Levine}.} \bibinfo{year}{2018}\natexlab{}.
\newblock \showarticletitle{Soft Actor-Critic: Off-Policy Maximum Entropy Deep
  Reinforcement Learning with a Stochastic Actor}.
\newblock \bibinfo{journal}{\emph{arXiv preprint arXiv:abs/1801.01290}}
  (\bibinfo{year}{2018}).
\newblock


\bibitem[\protect\citeauthoryear{Hasselt}{Hasselt}{2010}]%
        {hasselt2010}
\bibfield{author}{\bibinfo{person}{Hado~van Hasselt}.}
  \bibinfo{year}{2010}\natexlab{}.
\newblock \showarticletitle{Double Q-learning}. In
  \bibinfo{booktitle}{\emph{Proceedings of the 23rd International Conference on
  Neural Information Processing Systems - Volume 2}}
  \emph{(\bibinfo{series}{NIPS'10})}. \bibinfo{publisher}{Curran Associates
  Inc.}, \bibinfo{address}{USA}, \bibinfo{pages}{2613--2621}.
\newblock
\urldef\tempurl%
\url{http://dl.acm.org/citation.cfm?id=2997046.2997187}
\showURL{%
\tempurl}


\bibitem[\protect\citeauthoryear{Hoenig, Milanes, Scaria, Phan, Bolas, and
  Ayanian}{Hoenig et~al\mbox{.}}{2015}]%
        {hoenig2015mixed}
\bibfield{author}{\bibinfo{person}{Wolfgang Hoenig}, \bibinfo{person}{Christina
  Milanes}, \bibinfo{person}{Lisa Scaria}, \bibinfo{person}{Thai Phan},
  \bibinfo{person}{Mark Bolas}, {and} \bibinfo{person}{Nora Ayanian}.}
  \bibinfo{year}{2015}\natexlab{}.
\newblock \showarticletitle{Mixed reality for robotics}. In
  \bibinfo{booktitle}{\emph{2015 IEEE/RSJ International Conference on
  Intelligent Robots and Systems (IROS)}}. IEEE, \bibinfo{pages}{5382--5387}.
\newblock


\bibitem[\protect\citeauthoryear{Hyldmar, He, and Prorok}{Hyldmar
  et~al\mbox{.}}{2019}]%
        {hyldmar_Fleet_2019}
\bibfield{author}{\bibinfo{person}{Nicholas Hyldmar}, \bibinfo{person}{Yijun
  He}, {and} \bibinfo{person}{Amanda Prorok}.} \bibinfo{year}{2019}\natexlab{}.
\newblock \showarticletitle{A {Fleet} of {Miniature} {Cars} for {Experiments}
  in {Cooperative} {Driving}}.
\newblock \bibinfo{journal}{\emph{IEEE International Conference Robotics and
  Automation (ICRA)}} (\bibinfo{year}{2019}).
\newblock
\urldef\tempurl%
\url{https://doi.org/10.17863/CAM.37116}
\showDOI{\tempurl}


\bibitem[\protect\citeauthoryear{James, Wohlhart, Kalakrishnan, Kalashnikov,
  Irpan, Ibarz, Levine, Hadsell, and Bousmalis}{James et~al\mbox{.}}{2019}]%
        {james2019sim}
\bibfield{author}{\bibinfo{person}{Stephen James}, \bibinfo{person}{Paul
  Wohlhart}, \bibinfo{person}{Mrinal Kalakrishnan}, \bibinfo{person}{Dmitry
  Kalashnikov}, \bibinfo{person}{Alex Irpan}, \bibinfo{person}{Julian Ibarz},
  \bibinfo{person}{Sergey Levine}, \bibinfo{person}{Raia Hadsell}, {and}
  \bibinfo{person}{Konstantinos Bousmalis}.} \bibinfo{year}{2019}\natexlab{}.
\newblock \showarticletitle{Sim-to-real via sim-to-sim: Data-efficient robotic
  grasping via randomized-to-canonical adaptation networks}. In
  \bibinfo{booktitle}{\emph{Proceedings of the IEEE Conference on Computer
  Vision and Pattern Recognition}}. \bibinfo{pages}{12627--12637}.
\newblock


\bibitem[\protect\citeauthoryear{Kesting, Treiber, and Helbing}{Kesting
  et~al\mbox{.}}{2007}]%
        {kesting2007general}
\bibfield{author}{\bibinfo{person}{Arne Kesting}, \bibinfo{person}{Martin
  Treiber}, {and} \bibinfo{person}{Dirk Helbing}.}
  \bibinfo{year}{2007}\natexlab{}.
\newblock \showarticletitle{General Lane-Changing Model MOBIL for Car-Following
  Models}.
\newblock \bibinfo{journal}{\emph{Transportation Research Record}}
  \bibinfo{volume}{1999}, \bibinfo{number}{1} (\bibinfo{year}{2007}),
  \bibinfo{pages}{86--94}.
\newblock
\urldef\tempurl%
\url{https://doi.org/10.3141/1999-10}
\showDOI{\tempurl}


\bibitem[\protect\citeauthoryear{Khan, Zhang, Li, Wu, Schlotfeldt, Tang,
  Ribeiro, Bastani, and Kumar}{Khan et~al\mbox{.}}{2019}]%
        {khan2019learning}
\bibfield{author}{\bibinfo{person}{Arbaaz Khan}, \bibinfo{person}{Chi Zhang},
  \bibinfo{person}{Shuo Li}, \bibinfo{person}{Jiayue Wu},
  \bibinfo{person}{Brent Schlotfeldt}, \bibinfo{person}{Sarah~Y Tang},
  \bibinfo{person}{Alejandro Ribeiro}, \bibinfo{person}{Osbert Bastani}, {and}
  \bibinfo{person}{Vijay Kumar}.} \bibinfo{year}{2019}\natexlab{}.
\newblock \showarticletitle{Learning safe unlabeled multi-robot planning with
  motion constraints}.
\newblock \bibinfo{journal}{\emph{arXiv preprint arXiv:1907.05300}}
  (\bibinfo{year}{2019}).
\newblock


\bibitem[\protect\citeauthoryear{Kuderer, Gulati, and Burgard}{Kuderer
  et~al\mbox{.}}{2015}]%
        {kuderer2015learning}
\bibfield{author}{\bibinfo{person}{Markus Kuderer}, \bibinfo{person}{Shilpa
  Gulati}, {and} \bibinfo{person}{Wolfram Burgard}.}
  \bibinfo{year}{2015}\natexlab{}.
\newblock \showarticletitle{Learning driving styles for autonomous vehicles
  from demonstration}. In \bibinfo{booktitle}{\emph{2015 IEEE International
  Conference on Robotics and Automation (ICRA)}}. IEEE,
  \bibinfo{pages}{2641--2646}.
\newblock


\bibitem[\protect\citeauthoryear{Lowrey, Kolev, Dao, Rajeswaran, and
  Todorov}{Lowrey et~al\mbox{.}}{2018}]%
        {lowrey2018reinforcement}
\bibfield{author}{\bibinfo{person}{Kendall Lowrey}, \bibinfo{person}{Svetoslav
  Kolev}, \bibinfo{person}{Jeremy Dao}, \bibinfo{person}{Aravind Rajeswaran},
  {and} \bibinfo{person}{Emanuel Todorov}.} \bibinfo{year}{2018}\natexlab{}.
\newblock \showarticletitle{Reinforcement learning for non-prehensile
  manipulation: Transfer from simulation to physical system}. In
  \bibinfo{booktitle}{\emph{2018 IEEE International Conference on Simulation,
  Modeling, and Programming for Autonomous Robots (SIMPAR)}}. IEEE,
  \bibinfo{pages}{35--42}.
\newblock


\bibitem[\protect\citeauthoryear{Miglino, Lund, and Nolfi}{Miglino
  et~al\mbox{.}}{1995}]%
        {miglino1995evolving}
\bibfield{author}{\bibinfo{person}{Orazio Miglino},
  \bibinfo{person}{Henrik~Hautop Lund}, {and} \bibinfo{person}{Stefano Nolfi}.}
  \bibinfo{year}{1995}\natexlab{}.
\newblock \showarticletitle{Evolving mobile robots in simulated and real
  environments}.
\newblock \bibinfo{journal}{\emph{Artificial life}} \bibinfo{volume}{2},
  \bibinfo{number}{4} (\bibinfo{year}{1995}), \bibinfo{pages}{417--434}.
\newblock


\bibitem[\protect\citeauthoryear{Mnih, Badia, Mirza, Graves, Lillicrap, Harley,
  Silver, and Kavukcuoglu}{Mnih et~al\mbox{.}}{2016}]%
        {a3c}
\bibfield{author}{\bibinfo{person}{Volodymyr Mnih},
  \bibinfo{person}{Adri{\`{a}}~Puigdom{\`{e}}nech Badia},
  \bibinfo{person}{Mehdi Mirza}, \bibinfo{person}{Alex Graves},
  \bibinfo{person}{Timothy~P. Lillicrap}, \bibinfo{person}{Tim Harley},
  \bibinfo{person}{David Silver}, {and} \bibinfo{person}{Koray Kavukcuoglu}.}
  \bibinfo{year}{2016}\natexlab{}.
\newblock \showarticletitle{Asynchronous Methods for Deep Reinforcement
  Learning}.
\newblock \bibinfo{journal}{\emph{arXiv preprint arXiv:1602.01783}}
  (\bibinfo{year}{2016}).
\newblock


\bibitem[\protect\citeauthoryear{Mohammadi, Zamani, Kerzel, and
  Wermter}{Mohammadi et~al\mbox{.}}{2019}]%
        {mohammadi2019mixed}
\bibfield{author}{\bibinfo{person}{Hadi~Beik Mohammadi},
  \bibinfo{person}{Mohammad~Ali Zamani}, \bibinfo{person}{Matthias Kerzel},
  {and} \bibinfo{person}{Stefan Wermter}.} \bibinfo{year}{2019}\natexlab{}.
\newblock \showarticletitle{Mixed-Reality Deep Reinforcement Learning for a
  Reach-to-grasp Task}. In \bibinfo{booktitle}{\emph{International Conference
  on Artificial Neural Networks}}. Springer, \bibinfo{pages}{611--623}.
\newblock


\bibitem[\protect\citeauthoryear{Molchanov, Chen, H\"{o}nig, Preiss, Ayanian, and
  Sukhatme}{Molchanov et~al\mbox{.}}{2019}]%
        {molchanovSimto2019}
\bibfield{author}{\bibinfo{person}{Artem Molchanov}, \bibinfo{person}{Tao
  Chen}, \bibinfo{person}{Wolfgang H\"{o}nig}, \bibinfo{person}{James~A. Preiss},
  \bibinfo{person}{Nora Ayanian}, {and} \bibinfo{person}{Gaurav~S. Sukhatme}.}
  \bibinfo{year}{2019}\natexlab{}.
\newblock \showarticletitle{Sim-to-({Multi})-{Real}: {Transfer} of
  {Low}-{Level} {Robust} {Control} {Policies} to {Multiple} {Quadrotors}}.
\newblock \bibinfo{journal}{\emph{arXiv:1903.04628 [cs]}}
  (\bibinfo{date}{March} \bibinfo{year}{2019}).
\newblock
\urldef\tempurl%
\url{http://arxiv.org/abs/1903.04628}
\showURL{%
\tempurl}
\newblock
\shownote{arXiv: 1903.04628.}


\bibitem[\protect\citeauthoryear{Muratore, Treede, Gienger, and
  Peters}{Muratore et~al\mbox{.}}{2018}]%
        {muratore2018domain}
\bibfield{author}{\bibinfo{person}{Fabio Muratore}, \bibinfo{person}{Felix
  Treede}, \bibinfo{person}{Michael Gienger}, {and} \bibinfo{person}{Jan
  Peters}.} \bibinfo{year}{2018}\natexlab{}.
\newblock \showarticletitle{Domain randomization for simulation-based policy
  optimization with transferability assessment}. In
  \bibinfo{booktitle}{\emph{Conference on Robot Learning}}.
  \bibinfo{pages}{700--713}.
\newblock


\bibitem[\protect\citeauthoryear{Nagabandi, Clavera, Liu, Fearing, Abbeel,
  Levine, and Finn}{Nagabandi et~al\mbox{.}}{2019}]%
        {nagabandi_Learning_2019}
\bibfield{author}{\bibinfo{person}{Anusha Nagabandi}, \bibinfo{person}{Ignasi
  Clavera}, \bibinfo{person}{Simin Liu}, \bibinfo{person}{Ronald~S. Fearing},
  \bibinfo{person}{Pieter Abbeel}, \bibinfo{person}{Sergey Levine}, {and}
  \bibinfo{person}{Chelsea Finn}.} \bibinfo{year}{2019}\natexlab{}.
\newblock \showarticletitle{Learning to {Adapt} in {Dynamic}, {Real}-{World}
  {Environments} {Through} {Meta}-{Reinforcement} {Learning}}.
\newblock \bibinfo{journal}{\emph{arXiv:1803.11347 [cs, stat]}}
  (\bibinfo{date}{Feb.} \bibinfo{year}{2019}).
\newblock
\urldef\tempurl%
\url{http://arxiv.org/abs/1803.11347}
\showURL{%
\tempurl}
\newblock
\shownote{arXiv: 1803.11347.}


\bibitem[\protect\citeauthoryear{Pan, You, Wang, and Lu}{Pan
  et~al\mbox{.}}{2017}]%
        {pan2017virtual}
\bibfield{author}{\bibinfo{person}{Xinlei Pan}, \bibinfo{person}{Yurong You},
  \bibinfo{person}{Ziyan Wang}, {and} \bibinfo{person}{Cewu Lu}.}
  \bibinfo{year}{2017}\natexlab{}.
\newblock \showarticletitle{Virtual to real reinforcement learning for
  autonomous driving}.
\newblock \bibinfo{journal}{\emph{arXiv preprint arXiv:1704.03952}}
  (\bibinfo{year}{2017}).
\newblock


\bibitem[\protect\citeauthoryear{Peng, Andrychowicz, Zaremba, and Abbeel}{Peng
  et~al\mbox{.}}{2018}]%
        {peng2018sim}
\bibfield{author}{\bibinfo{person}{Xue~Bin Peng}, \bibinfo{person}{Marcin
  Andrychowicz}, \bibinfo{person}{Wojciech Zaremba}, {and}
  \bibinfo{person}{Pieter Abbeel}.} \bibinfo{year}{2018}\natexlab{}.
\newblock \showarticletitle{Sim-to-real transfer of robotic control with
  dynamics randomization}. In \bibinfo{booktitle}{\emph{2018 IEEE International
  Conference on Robotics and Automation (ICRA)}}. IEEE, \bibinfo{pages}{1--8}.
\newblock


\bibitem[\protect\citeauthoryear{Schulman, Wolski, Dhariwal, Radford, and
  Klimov}{Schulman et~al\mbox{.}}{2017}]%
        {ppo}
\bibfield{author}{\bibinfo{person}{John Schulman}, \bibinfo{person}{Filip
  Wolski}, \bibinfo{person}{Prafulla Dhariwal}, \bibinfo{person}{Alec Radford},
  {and} \bibinfo{person}{Oleg Klimov}.} \bibinfo{year}{2017}\natexlab{}.
\newblock \showarticletitle{Proximal Policy Optimization Algorithms}.
\newblock \bibinfo{journal}{\emph{arXiv preprint arXiv:1707.06347}}
  (\bibinfo{year}{2017}).
\newblock


\bibitem[\protect\citeauthoryear{Shah, Dey, Lovett, and Kapoor}{Shah
  et~al\mbox{.}}{2018}]%
        {shah2018airsim}
\bibfield{author}{\bibinfo{person}{Shital Shah}, \bibinfo{person}{Debadeepta
  Dey}, \bibinfo{person}{Chris Lovett}, {and} \bibinfo{person}{Ashish Kapoor}.}
  \bibinfo{year}{2018}\natexlab{}.
\newblock \showarticletitle{Airsim: High-fidelity visual and physical
  simulation for autonomous vehicles}. In \bibinfo{booktitle}{\emph{Field and
  service robotics}}. Springer, \bibinfo{pages}{621--635}.
\newblock


\bibitem[\protect\citeauthoryear{Shalev-Shwartz, Shammah, and
  Shashua}{Shalev-Shwartz et~al\mbox{.}}{2016}]%
        {shalev2016safe}
\bibfield{author}{\bibinfo{person}{Shai Shalev-Shwartz},
  \bibinfo{person}{Shaked Shammah}, {and} \bibinfo{person}{Amnon Shashua}.}
  \bibinfo{year}{2016}\natexlab{}.
\newblock \showarticletitle{Safe, multi-agent, reinforcement learning for
  autonomous driving}.
\newblock \bibinfo{journal}{\emph{arXiv preprint arXiv:1610.03295}}
  (\bibinfo{year}{2016}).
\newblock


\bibitem[\protect\citeauthoryear{Stilman, Michel, Chestnutt, Nishiwaki, Kagami,
  and Kuffner}{Stilman et~al\mbox{.}}{2005}]%
        {stilman2005augmented}
\bibfield{author}{\bibinfo{person}{Michael Stilman}, \bibinfo{person}{Philipp
  Michel}, \bibinfo{person}{Joel Chestnutt}, \bibinfo{person}{Koichi
  Nishiwaki}, \bibinfo{person}{Satoshi Kagami}, {and} \bibinfo{person}{James
  Kuffner}.} \bibinfo{year}{2005}\natexlab{}.
\newblock \showarticletitle{Augmented reality for robot development and
  experimentation}.
\newblock \bibinfo{journal}{\emph{Robotics Institute, Carnegie Mellon
  University, Pittsburgh, PA, Tech. Rep. CMU-RI-TR-05-55}} \bibinfo{volume}{2},
  \bibinfo{number}{3} (\bibinfo{year}{2005}).
\newblock


\bibitem[\protect\citeauthoryear{Sutton and Barto}{Sutton and Barto}{2011}]%
        {sutton2011}
\bibfield{author}{\bibinfo{person}{Richard~S Sutton} {and}
  \bibinfo{person}{Andrew~G Barto}.} \bibinfo{year}{2011}\natexlab{}.
\newblock \showarticletitle{Reinforcement learning: An introduction}.
\newblock  (\bibinfo{year}{2011}).
\newblock


\bibitem[\protect\citeauthoryear{Tan, Zhang, Coumans, Iscen, Bai, Hafner,
  Bohez, and Vanhoucke}{Tan et~al\mbox{.}}{2018}]%
        {tan2018sim}
\bibfield{author}{\bibinfo{person}{Jie Tan}, \bibinfo{person}{Tingnan Zhang},
  \bibinfo{person}{Erwin Coumans}, \bibinfo{person}{Atil Iscen},
  \bibinfo{person}{Yunfei Bai}, \bibinfo{person}{Danijar Hafner},
  \bibinfo{person}{Steven Bohez}, {and} \bibinfo{person}{Vincent Vanhoucke}.}
  \bibinfo{year}{2018}\natexlab{}.
\newblock \showarticletitle{Sim-to-real: Learning agile locomotion for
  quadruped robots}.
\newblock \bibinfo{journal}{\emph{arXiv preprint arXiv:1804.10332}}
  (\bibinfo{year}{2018}).
\newblock


\bibitem[\protect\citeauthoryear{Tobin, Fong, Ray, Schneider, Zaremba, and
  Abbeel}{Tobin et~al\mbox{.}}{2017}]%
        {tobin_Domain_2017}
\bibfield{author}{\bibinfo{person}{J. Tobin}, \bibinfo{person}{R. Fong},
  \bibinfo{person}{A. Ray}, \bibinfo{person}{J. Schneider}, \bibinfo{person}{W.
  Zaremba}, {and} \bibinfo{person}{P. Abbeel}.}
  \bibinfo{year}{2017}\natexlab{}.
\newblock \showarticletitle{Domain randomization for transferring deep neural
  networks from simulation to the real world}. In
  \bibinfo{booktitle}{\emph{2017 {IEEE}/{RSJ} {International} {Conference} on
  {Intelligent} {Robots} and {Systems} ({IROS})}}. \bibinfo{pages}{23--30}.
\newblock
\urldef\tempurl%
\url{https://doi.org/10.1109/IROS.2017.8202133}
\showDOI{\tempurl}


\bibitem[\protect\citeauthoryear{Treiber, Hennecke, and Helbing}{Treiber
  et~al\mbox{.}}{2000}]%
        {treiber2000congested}
\bibfield{author}{\bibinfo{person}{Martin Treiber}, \bibinfo{person}{Ansgar
  Hennecke}, {and} \bibinfo{person}{Dirk Helbing}.}
  \bibinfo{year}{2000}\natexlab{}.
\newblock \showarticletitle{Congested traffic states in empirical observations
  and microscopic simulations}.
\newblock \bibinfo{journal}{\emph{Phys. Rev. E}}  \bibinfo{volume}{62}
  (\bibinfo{date}{Aug} \bibinfo{year}{2000}), \bibinfo{pages}{1805--1824}.
\newblock
Issue 2.
\urldef\tempurl%
\url{https://doi.org/10.1103/PhysRevE.62.1805}
\showDOI{\tempurl}


\bibitem[\protect\citeauthoryear{Williams, Szafir, Chakraborti, and
  Ben~Amor}{Williams et~al\mbox{.}}{2018}]%
        {williams2018virtual}
\bibfield{author}{\bibinfo{person}{Tom Williams}, \bibinfo{person}{Daniel
  Szafir}, \bibinfo{person}{Tathagata Chakraborti}, {and} \bibinfo{person}{Heni
  Ben~Amor}.} \bibinfo{year}{2018}\natexlab{}.
\newblock \showarticletitle{Virtual, augmented, and mixed reality for
  human-robot interaction}. In \bibinfo{booktitle}{\emph{Companion of the 2018
  ACM/IEEE International Conference on Human-Robot Interaction}}. ACM,
  \bibinfo{pages}{403--404}.
\newblock


\end{thebibliography}
\end{document}